\documentclass[twoside,11pt]{article}
\usepackage{jair, theapa, rawfonts,amsmath,graphicx,amsfonts}

\ShortHeadings{Neural Network Plasticity and Loss Sharpness}
{Koster \& Kukla}
\firstpageno{1}
\newcommand{\norm}[1]{\left\lVert#1\right\rVert}

\begin{document}

\title{Neural Network Plasticity and Loss Sharpness}

\author{
% \name Steven Minton \email minton@ptolemy.arc.nasa.gov \\
       % \name Max Koster \email mok25@cornell.edu \\
       \name Max Koster \email mok25@cornell.edu \\
       \addr Cornell University \\ 
       Department of Computer Science\\
       Ithaca, NY 14850, USA
       \AND
       \name Jude Kukla \email cjk264@cornell.edu \\
       \addr Cornell University \\ 
       Department of Computer Science\\
       Ithaca, NY 14850, USA
}
% For research notes, remove the comment character in the line below.
% \researchnote

\maketitle

\begin{abstract}
In recent years, continual learning—a prediction setting in which the problem environment may evolve over time—has become an increasingly popular research field due to the framework's gearing towards complex, non-stationary objectives.  Learning such objectives requires plasticity, or the ability of a neural network to adapt its predictions to a different task. Recent findings indicate that plasticity loss on new tasks is highly related to loss landscape sharpness in non-stationary RL frameworks. 
We explore the usage of sharpness regularization techniques, which seek out smooth minima and have been touted for their generalization capabilities in vanilla prediction settings, in efforts to combat plasticity loss. Our findings indicate that such techniques have no significant effect on reducing plasticity loss.  
\end{abstract}

\section{Introduction}
\label{Introduction}

The traditional supervised learning setup is that in which a model learns a function which is invariant over time and model training occurs once. However, in various learning domains, the function may be non-stationary. In a discrete-time setting, a non-stationary objective consists of training a neural net on some series of related, yet distinct tasks. 

This is referred to this as a continual learning setting, where the model faces a changing stream of data to learn \cite{wang2023comprehensive}. Consider an email spam classifier: the vocabulary of spam emails changes over time, and would thus require re-training from scratch periodically in a train once setting. Instead, in a continual learning setting, the model would continue training and simply adapt to changes in spam trends.  

Naturally, the central objective of a continual learning model is to maintain performance throughout various tasks. The decreased ability of a neural network to perform accurately on later tasks is described as a loss of plasticity. The loss of plasticity in current deep learning models was observed as early as the first deep networks and continues to persist in modern deep learning \cite{dohare2023loss}.

\begin{figure}[ht]
    \centering
    \includegraphics[width=6cm]{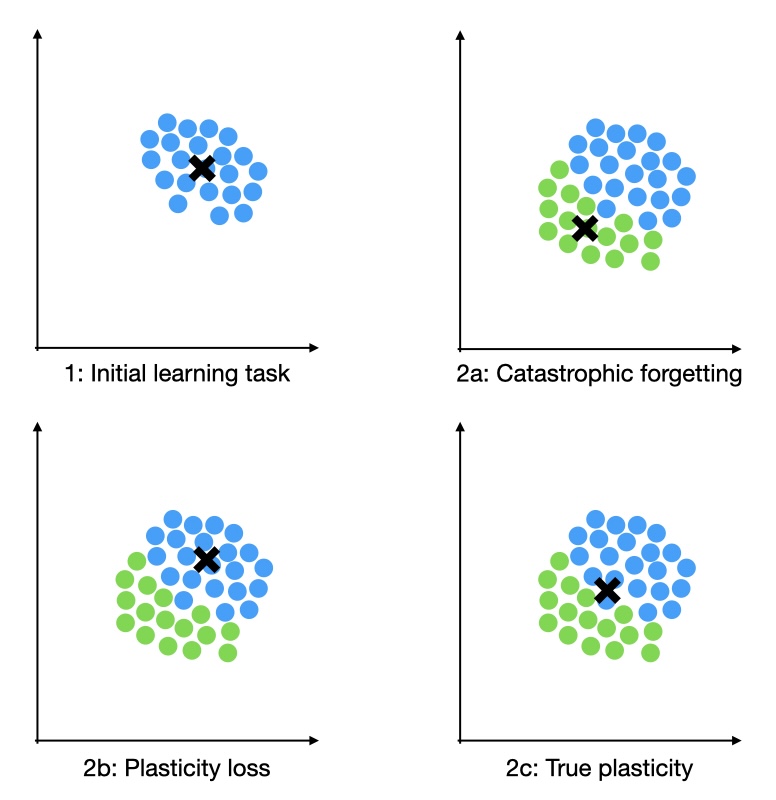}
    \caption{ Blue and green regions represents set of parameters $\theta$ with high performance for two tasks $A$ and $B$, respectively. 1: Model learns parameter $\textbf{x}$ for task $A$ \& achieves high accuracy 2a: Following the introduction of $B$, model learns some $\textbf{x}$ that produces high performance on $B$  with poor performance on $A$. 2b: Model maintains performance on $A$ but obtains poor performance on $B$.  2c: Model learns $\textbf{x}$ with good performance on both $A$ and $B$.}
     \label{fig:plasticity}
\end{figure}

Figure~\ref{fig:plasticity} presents a visual representation of different outcomes in the process of learning and adapting to new tasks, as demonstrated in a neural network or a learning system. Each colored region represents regions with low training accuracy. The top left quadrant (1) shows the state of the system after an initial learning task. The top right quadrant (2a) depicts ``catastrophic forgetting,'' where the introduction of a new task causes the system to shift its focus entirely to the new information, losing the previously learned information. The bottom left quadrant (2b) illustrates ``plasticity loss,'' where the system retains the original task but fails to adapt to the new task. Lastly, the bottom right quadrant (2c) represents ``true plasticity,'' where the system successfully performs on the new \& old task.
It is important to note the relevance of continual learning to deep reinforcement learning problems. Consider the Deep Q-network algorithm, in which the optimal policy is not known and instead we must calculate a bootstrapped version for each gradient step \cite{mnih2013playing}. Thus, our function is clearly non-stationary. However, we do not perform empirical studies using an deep reinforcement learning example due to its inherent complexities; instead, we investigate loss of plasticity in the more generalized continual learning setting.

Recent works have focused on investigating causes of plasticity loss and to possibly discover various techniques to counteract such loss. One area of interest is the network’s loss environment and more specifically the loss sharpness, which can be quantified through the  maximal eigenvalue of the loss’s Hessian matrix. Lyle et al. found that the maximal eigenvalue increases over tasks; that is, sharpness of minima increases as more tasks are presented to the network \cite{lyle2023understanding}. 

In this paper, we introduce the usage of sharpness regularization techniques to prevent plasticity loss in non-stationary functions. Such techniques bias networks to find 'flatter' minima. These methods have been found to improve generalization ability on traditional supervised learning tasks. Our usage of sharpness regularization techniques is clearly motivated by the results of \cite{lyle2023understanding}.

Beyond these recent empirical results, it is also reasonable to posit that if a minima of task A has greater generalization ability, it will be more transferable to task B; thus, we should seek such minima.  
\section{Previous Work}
\label{network}
\subsection{Flat Minima}
The relationship between flat minima and generalization ability has become well-examined in recent years.       
The actual mathematical formulation of what a ‘flat minima’ is can differ in various literature, but it is unequivocally thought of as a region where loss remains relatively constant. Here, we present a simple approach via Lipschitz continuity. Recall that a function $f: \mathbb{R}^n \rightarrow \mathbb{R}^n $ is Lipschitz continuous everywhere if $\forall x_1, x_2 \in \mathbb{R}^n$ there exists a Lipschitz constant $K$ such that: $$ \norm{f(x_1) - f(x_2)}_2 \leq K \cdot \norm{x_1 - x_2}_2 $$
Selection of the Lipschitz constant is typically the minimum such value for which the equality holds. Also, Lipschitz continuity can be described over some local neighborhood $V_\epsilon(x)$. In this case, the value $K$ is an upper bound for the perturbation or change in $f(x)$ in some epsilon neighborhood of $x$. This is a clear framework to quantify the ‘sharpness’ of a local minima. 

Recently, there has been progress in developing optimization schema which explicitly seek out flat minima. Sharpness-Aware Minimization (SAM) minimizes loss sharpness by penalizing the loss in some $\epsilon$ neighborhood of $w$ \cite{foret2020sharpness}. This is formalized as the following minimization: 
$$ \min_{\mathbf{w}} L_\mathcal{S}^{SAM}(w) + \lambda \norm{w}_2
$$
where 
$$L_\mathcal{S}^{SAM}(w) = \max_{\norm{\epsilon}_2 \leq \rho } L_\mathcal{S}(\mathbf{w} + \epsilon)$$
% This is a minimax optimization problem and thus solved via linear programming techniques. 
Gradient Norm Penalty (GNP) instead explicitly penalizes the gradient of the loss $L_\mathcal{S}(\mathbf{w})$ \cite{zhao2022penalizing}: 
$$L(\mathbf{w}) = L_\mathcal{S}(\mathbf{w}) + \lambda\norm{\nabla L_\mathcal{S}(\mathbf{w})}_2$$
The reasoning follows from the previously presented Lipshitz definition. We first recall the Mean Value Theorem from real analysis: if $f(x)$ is differentiable on $[a,b]$, then there exists $c \in [a,b]$ such that  $\nabla f(c)(a - b) = f(a) - f(b) $

% $(c) = \frac{f(a) - f(b)}{a-b}$.

Now consider an arbitrary minimum $\mathbf{w}$ and some neighborhood $V_\epsilon(\mathbf{w})$. 
Then $\exists \zeta \in V_\epsilon(\mathbf{w})$ such that $\forall w_i \in V_\epsilon(\mathbf{w})$, $$\norm{\nabla(\zeta)(\mathbf{w} - w_i)}_2 = \norm{f(\mathbf{w}) - f(w_i)}_2 $$ 
Then it follows from the Cauchy–Schwarz inequality that 
$$ \norm{f(\mathbf{w}) - f(w_i)}_2  \leq \norm{\nabla(\zeta)}_2\norm{\mathbf{w} - w_i}_2$$ 
As $\epsilon {\rightarrow} 0$, $\zeta {\rightarrow} \mathbf{w}$, so we approximate $\norm{\nabla(\zeta)}_2$ as $\norm{\nabla(\mathbf{w})}_2$. Then it follows that decreasing $\norm{\nabla(\mathbf{w})}_2$ is equivalent to lowering the Lipschitz constant $K$ in a small $\epsilon$-neighborhood of $\mathbf{w}$. Thus, penalization of the gradient norm is a penalization of the sharpness of a minima. These regularization techniques improved upon  performance benchmarks of image classification tasks, with SAM providing new state-of-the-art accuracy on CIFAR-100 that GNP subsequently improved upon.
\subsection{Plasticity}
The factors which cause plasticity of neural networks are still poorly understood; this is an active research area. Some explored potential causes include feature inactivity, increased weight norm, and low rank of features \cite{lyle2023understanding}. Considerable attention is also given to a multitude of proposed mechanisms to prevent plasticity loss, including layer normalization, weight decay, Shrink and Perturb and spectral normalization. 
\newline Past empirical work have introduced spectral normalization techniques to deep reinforcement learning problems such as Atari. Said techniques similarly exploit the relation between the gradient norm and Lipshitz continuity. However, rather than modifying the loss function, they use a typical loss function but project parameters into a subspace with a desired spectral radius after each gradient step. \cite{specnormRL} We also note that said technique was solely applied to specific network layers rather than the whole network. 
To our knowledge, our work represents the first attempt to directly apply neural network sharpness regularization techniques to the continual learning setting. 
 % Since gradient descent requires us to take the derivative with respect to the loss function (which includes the gradient), a naive implementation would require explicit computation of the Hessian, which is quite expensive. Instead, we []. 
\subsection{Continual Learning}
Continual Learning is a very broad problem setting with many subcategories. We formalize this notation by describing all data \(D  = \{(x_1, y_1), (x_2, y_2), \ldots, (x_n, y_n)\} \subset \mathbb{R}^d \times Y\) with the total set of labels $y \in \mathcal{Y}$. Similar to \cite{wang2023comprehensive}, we define  an incoming batch $b$ of  a prediction class as \(D_{b} = \{X_{ b}, Y_{ b}\}\), where \(X_{b}\) is the input data, \(Y_{b}\) is the data label  and \(b \in B_t\) is the batch index.  
%The entire dataset \(D = \bigcup D_{b} \)
% Some times there may be a task identity defined as proposed in denoting (\(T\) and \(B_t\) denote their space, respectively).
Though many various problem settings are discussed in literature, we now expand on two particularly popular and well-known ones: Class-Incremental \& Domain-Incremental Learning.
\subsubsection{Class-Incremental Learning}
Class-Incremental Learning is a classification problem setting in which additional classes are incrementally added. That is, each new batch $D_b$ introduces or removes some class.  each $D_b$ is composed of points drawn  from some specific subset $S \subseteq \mathcal{Y}$, or alternatively  \(D_b = \{(x_1, y_1), (x_2, y_2), \ldots, (x_n, y_n)\} \subset \mathbb{R}^d \times S\).
% . If we have some prior distribution \(D_{b} = \{X_{ b}, Y_{ b}\} = \bigcup_{ } \{X_{i_b}, Y_{i_b}\}\) given $f: \mathbb{R}^n \rightarrow \mathbb{R}^k $ 
\begin{figure*}[ht]
    \centering
    \includegraphics[width=\textwidth]{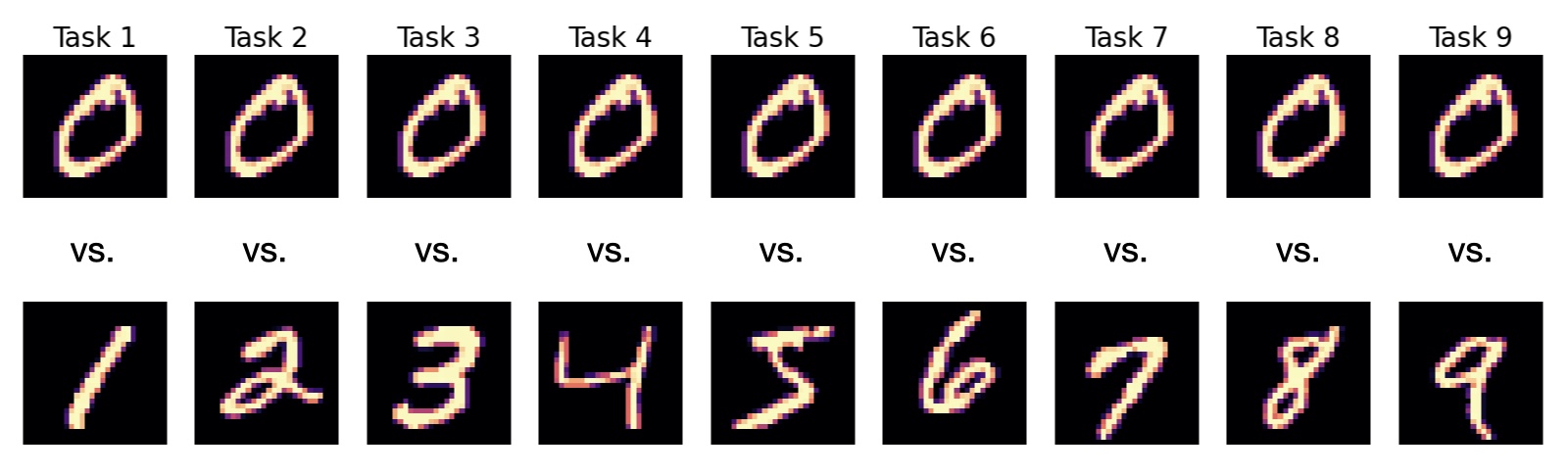}
    \caption{Nine class-incremental learning tasks}
    \label{fig:class-task}
\end{figure*}

\subsubsection{Domain-Incremental Learning}
Domain-Incremental Learning is a classification problem setting in which the classes stay the same, but the underlying distribution of a class may shift. That is, for each additional batch we still have \(D_b \subset \mathbb{R}^d \times Y\), but each batch is drawn from distinct distributions $P_b(x_i,y_i)$.
Our work aims to explore both of these types of continual learning.
\begin{figure*}[ht]
    \centering
    \includegraphics[width=\textwidth]{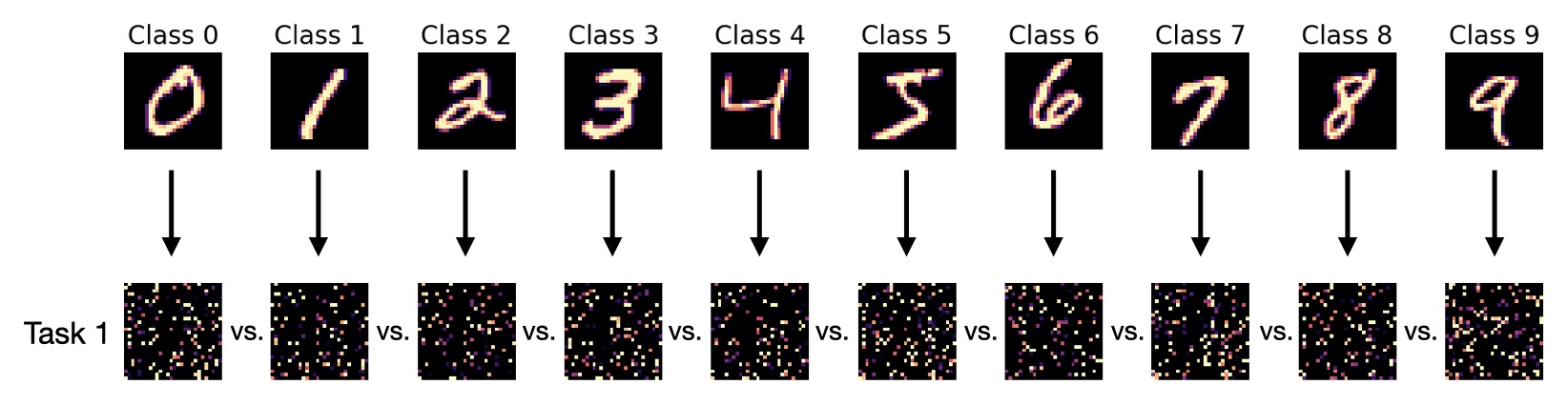}
    \caption{One domain-incremental learning task}
    \label{fig:domain-task}
\end{figure*}

\label{results}

\section{Experimental Setup}
\subsection{Data}
The MNIST database is a large set handwritten digits that is widely used in training and testing machine learning systems. It contains 60,000 training images and 10,000 testing images. Each image is \(28\times 28\), where each pixel can take on a value from 0 (black) to 1 (white).

In order to create a domain-incremental learning problem, we took the original version of this dataset and randomly permuted the pixels (\(10^{1930}\) possibilities) to create a single learning ``task,'' preserving the integrity of both the training and test sets with the creation of each task. Here, a randomly permuted dataset is a task (Figure~\ref{fig:domain-task}).

In order to create a class-incremental learning problem, we took the original version of the dataset, and extracted pairwise class data, taking care to maintain the 6:1 train-to-test ratio. A ``task'' here might be distinguishing the digit 1 from the digit 2, or the digit 7 from the digit 3. That is, a dataset containing two classes is a task. Figure \ref{fig:class-task} displays 9 tasks.

\subsection{Classifier}
In both of these learning tasks, we replicate the techniques of Dohare et al. \cite{dohare2023loss} and used a simple feed-forward neural network. Our architecture consists of one \(28\times28\) input layer, followed by three fully connected linear layers with ReLU activation, and a either a 10-hot or 2-hot output input layer (for the domain-incremental and class-incremental learning problems, respectively).

It is worth noting that we use virtually the same architecture for both training settings in an effort to improve the comparability of our results and their potential for generalization. Here, there is no relevance in the addition of convolutional layers since in the domain-incremental learning problem all spatial information is lost after every task.
\subsection{Results}
We devise a set of 6 training settings in order to evaluate the performance of both SAM and GNP relative to SGD in both learning problems: SGD with \(\alpha=0.01,\) SGD with \(\alpha=0.001,\) SAM with \(\alpha=0.01,\) SAM with \(\alpha=0.001,\) GNP with \(\alpha=0.01,\) and GNP with \(\alpha=0.001.\) Note that we set \(\lambda=0.1\) for both the SAM and GNP models. We train our network 10 times each in these training settings and report the mean task-specific test accuracy after each task.

It is important to note that in both learning problems, t. This is integral to evaluating true plasticity in our network.
\subsection{Domain-Incremental Learning}
\begin{figure}[ht]
    \centering
    \includegraphics[width=8cm]{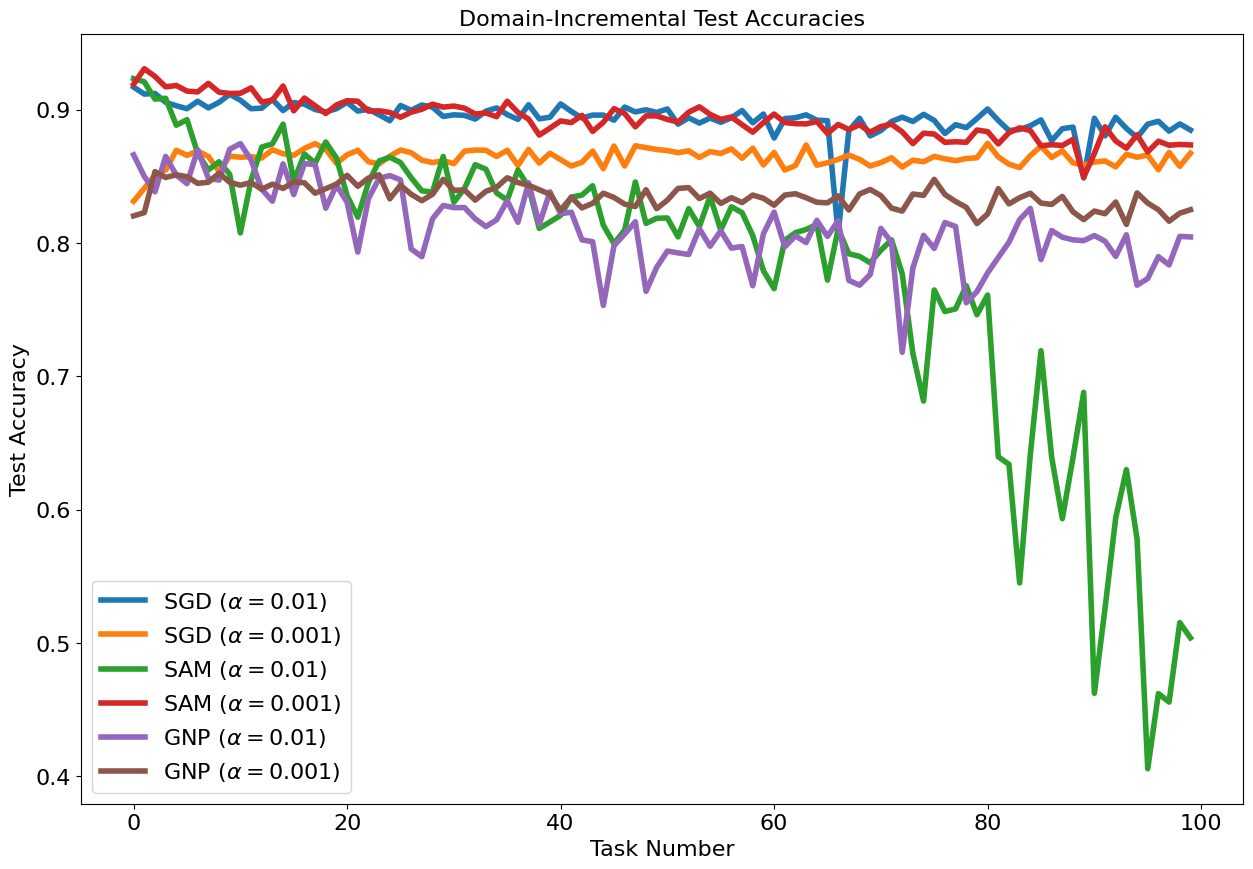}
    \caption{Task-specific accuracies in the domain-incremental learning problem under different loss minimization schema (10 runs)}
    \label{fig:domain-results}
\end{figure}

We plot the task-specific test accuracies (averaged over 10 runs) for each training setting below (Figure~\ref{fig:domain-results}) on 100 randomly selected domain-incremental learning tasks. Additionally, we report the average change in test accuracy over the 100-task training period (Table~\ref{tab:domain-table}).
\begin{table}[ht]
    \centering
    \begin{tabular}{c|c}
    Training Setting & Mean Per-Task Accuracy Change \\ \hline
       SGD (\(\alpha=0.01\)) &  \(-3.2\times10^{-4}\) \\
       SGD (\(\alpha=0.001\)) & \(+3.6\times10^{-4}\) \\
       SAM (\(\alpha=0.01\)) & \(-4.2\times10^{-3}\) \\
       SAM (\(\alpha=0.001\)) & \(-4.6\times10^{-4}\) \\
       GNP (\(\alpha=0.01\)) & \(-6.2\times10^{-4}\) \\
       GNP (\(\alpha=0.001\)) & \(+4.6\times10^{-5}\) 
    \end{tabular}
    \caption{Mean domain-incremental per-task accuracy changes under different loss minimization schema (10 runs)}
    \label{tab:domain-table}
\end{table}

\subsection{Class-Incremental Learning}
\begin{figure}[ht]
    \centering
    \includegraphics[width=8cm]{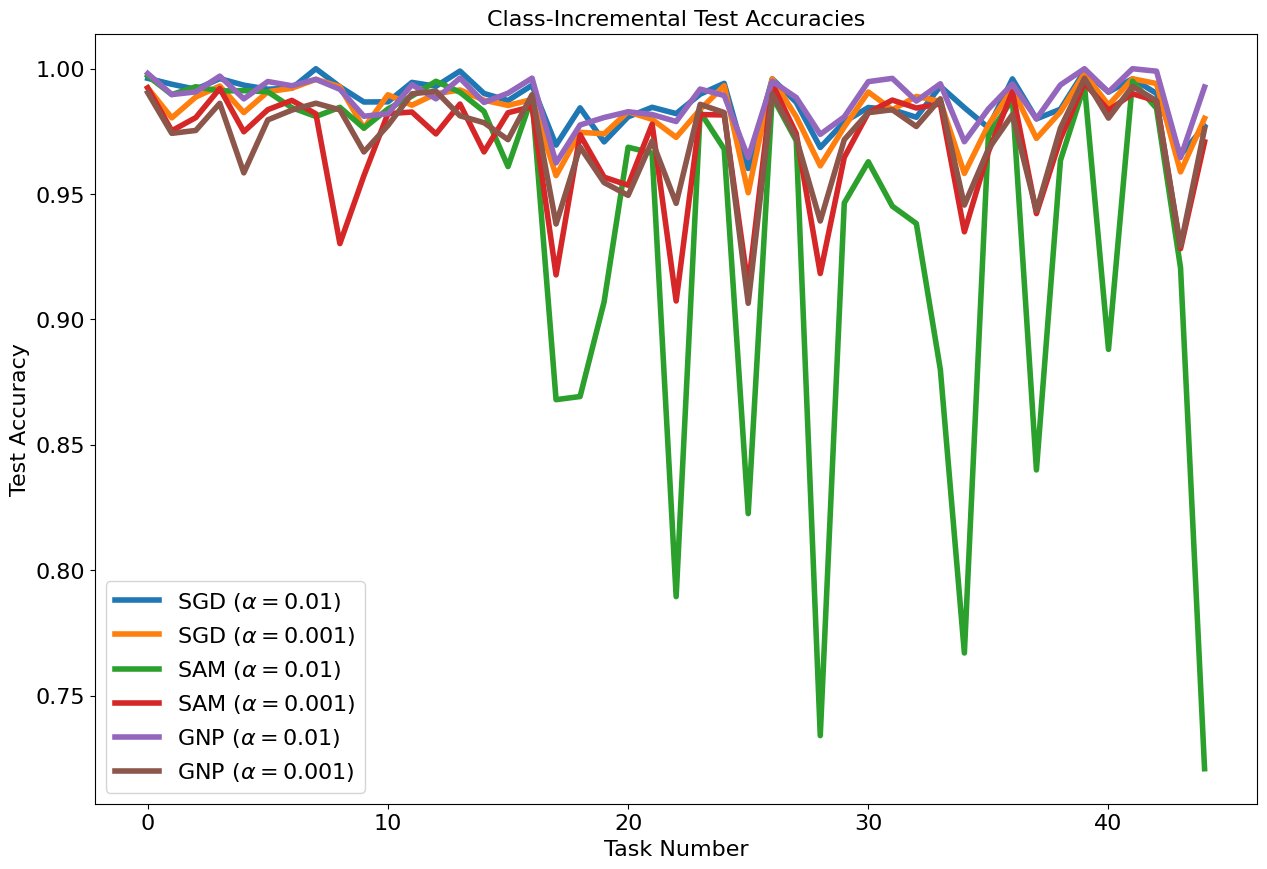}
    \caption{Task-specific accuracies in the class-incremental learning problem under different loss minimization schema (10 runs)}
    \label{fig:class-results}
\end{figure}

We again plot the task-specific test accuracies (averaged over 10 runs) for each training setting below (Figure~\ref{fig:class-results}) on the \({10\choose 2} =45\) possible tasks. Additionally, we report the average change in test accuracy over the 45-task training period (Table~\ref{tab:class-table}).

\begin{table}[ht]
    \centering
    \begin{tabular}{c|c}
    Training Setting & Mean Per-Task Accuracy Change \\ \hline
       SGD (\(\alpha=0.01\)) &  \(-4.2\times10^{-4}\) \\
       SGD (\(\alpha=0.001\)) & \(-2.7\times10^{-4}\) \\
       SAM (\(\alpha=0.01\)) & \(-6.1\times10^{-3}\) \\
       SAM (\(\alpha=0.001\)) & \(-4.8\times10^{-4}\) \\
       GNP (\(\alpha=0.01\)) & \(-1.2\times10^{-4}\) \\
       GNP (\(\alpha=0.001\)) & \(-3.2\times10^{-4}\) 
    \end{tabular}
    \caption{Mean task-incremental per-task accuracy changes under different loss minimization schema (10 runs)}
    \label{tab:class-table}
\end{table}

\section{Discussion}
Overall, our results demonstrate that sharpness regularization techniques like SAM and GNP does not decrease–and instead may worsen–plasticity loss in continual learning settings. 

In the domain-incremental learning experiments, all training methods showed to perform with similar accuracy over 100 tasks. SGD with \(\alpha=0.001\) performed the best in terms of maintaining performance, with a slight positive per-task accuracy change on average. SAM and GNP with higher learning rates showed noticeable declines in per-task accuracy over tasks.

% This aligns with recent findings that minima tend to get sharper over tasks, making it harder for networks to retain plasticity \cite{lyle2r023understanding}. The sharpness regularization techniques help counteract this to some extent.
% In the class-incremental learning setting, however, the differences between training methods were more apparent. SGD achieved stable per-task accuracy changes close to zero. In contrast, SAM showed significant declines in per-task accuracy, indicating substantial plasticity loss. GNP proved most effective at maintaining plasticity in this setting, with almost no change in per-task accuracy across the 45 class-incremental tasks.  

In the class-incremental learning setting, SGD achieved stable per-task accuracy changes close to zero. In contrast, SAM showed significant declines in per-task accuracy, indicating substantial plasticity loss. GNP proved most effective at maintaining plasticity in this setting, with almost no change in per-task accuracy across the 45 class-incremental tasks.  

% The fact that SAM and GNP provided a clearer benefit in the class-incremental versus domain-incremental case suggests the technique is best suited when there are distinct shifts between tasks requiring retention and integration of old knowledge with new knowledge. In the domain-incremental case, the permutations provide no transferable spatial patterns, requiring the network to relearn each task from scratch.
\subsection{Conclusion and Future Work}

Overall, our findings show that applying sharpness regularization alone to promote neural network plasticity will not effectively work. There are several possible explanations for this result. First, there is some question on the relevance of the Permuted MNIST dataset as an adequate example of continual learning \cite{farquhar2019robust}. Further empirical studies with different continual learning tasks can provide indication on whether this is the case. There is also the possibility that despite the empirical observations \cite{lyle2023understanding} of the relation between sharpness and plasticity, this is not causal and there exists some (currently) unknown confounding factor.

Key areas for future work include testing on more complex benchmark datasets as well as reinforcement learning problems with non-stationary reward functions over time. Adaptations to the sharpness regularization terms may also further improve continual learning performance—there is still much of the parameter space to be explored.

\acks{The authors wish to thank Yuta Saito and Thorsten Joachims for their assistance
and advice.  We also thank our anonymous reviewers for
their comments.   
}

\appendix
 
\vskip 0.2in
\bibliography{bib}
\bibliographystyle{theapa}

\end{document}